\documentclass[12pt]{article}
\usepackage{times}
\usepackage{graphicx}
\usepackage{mathtools}
\usepackage{amsmath}
\usepackage[displaymath,mathlines]{lineno}
\usepackage{color}
\usepackage[superscript]{cite}
\newenvironment{sciabstract}{%
	\begin{quote} \bf}
	{\end{quote}}

\title{\vspace{-20mm} Fluidic Fabric Muscle Sheets for \\ Wearable and Soft Robotics}

\author
{Mengjia Zhu,$^{1}$ Thanh Nho Do,$^{2}$ Elliot Hawkes,$^{3}$ Yon Visell$^{1,3\ast}$\\
	\\
	\normalsize{$^{1}$Media Arts and Technology Program, Department of Electrical and Computer Engineering,} \\ \normalsize{California NanoSystems Institute, and Center for Polymers and Organic Solids}, \\ \normalsize{University of California, Santa Barbara, USA}\\
	\normalsize{$^{2}$Graduate School of Biomedical Engineering, Faculty of Engineering},\\
	\normalsize{University of New South Wales, Sydney, Australia}\\
	\normalsize{$^{3}$Department of Mechanical Engineering, University of California, Santa Barbara, USA}\\
	\\
	\normalsize{$^\ast$To whom correspondence should be addressed; E-mail:  yonvisell@ucsb.edu.}
}
\date{}
\begin{document} 
	
	%	\baselineskip24pt
	\baselineskip13pt
	
	
	\maketitle

	
	\begin{sciabstract}
		Conformable robotic systems are attractive for applications in which they can be used to actuate structures with large surface areas, to provide forces through wearable garments, or to realize autonomous robotic systems. We present a new family of soft actuators that we refer to as Fluidic Fabric Muscle Sheets (FFMS). They are composite fabric structures that integrate fluidic transmissions based on arrays of elastic tubes. These sheet-like actuators can strain, squeeze, bend, and conform to hard or soft objects of arbitrary shapes or sizes, including the human body. We show how to design and fabricate FFMS actuators via facile apparel engineering methods, including computerized sewing techniques. Together, these determine the distributions of stresses and strains that can be generated by the FFMS. We present a simple mathematical model that proves effective for predicting their performance. FFMS can operate at frequencies of 5 Hertz or more,  achieve engineering strains exceeding 100\%, and  exert forces greater than 115 times their own weight. They can be safely used in intimate contact with the human body even when delivering stresses exceeding 10$^\text{6}$ Pascals.  We demonstrate their versatility for actuating a variety of bodies or structures, and in configurations that perform multi-axis actuation, including bending and shape change. As we also show,  FFMS can be used to exert forces on body tissues for wearable and biomedical applications. We demonstrate several potential use cases, including a miniature steerable robot, a glove for grasp assistance, garments for applying compression to the extremities, and devices for actuating small body regions or tissues via localized skin stretch. 
	\end{sciabstract}
	
	
	%\section*{Introduction}

	\section{Introduction}
	
	Emerging soft actuator technologies are enabling applications in robotics, healthcare, haptics, assistive technologies, and many other areas.  Such soft actuators can  interface with,  conform to,  exert forces upon, or generate shape changes in complex or compliant structures \cite{yap2016high, booth2018omniskins, rus2015design}. Wearable soft robotic devices interfaced with the human body may prove valuable for rehabilitation,  movement assistance, or virtual reality \cite{song2013soft, park2014design, koo2008development}.
	Soft actuators are also of interest for controlling motion in distributed or deformable structures. They can be used for tasks such as grasping, terrestrial locomotion, surgery, or underwater operation \cite{polygerinos2015soft, cotin1999real, serchi2013biomimetic}.  Such applications span systems of greatly varying length  scales, ranging from millimeter-scale biomedical robots to large, deployable structures \cite{kwon2008biomimetic, wang2016deployable}.  
	
	Biological systems provide a rich source of information to guide the  design of  soft robots \cite{kim2013soft}.  The motile capabilities of animals are enabled by composite systems of muscle, connective, and other tissues. The forces and motions they can produce  depend on the properties of individual muscle fibers, the  arrangement of fibers, and the muscle morphology and attachments.  Muscle morphologies vary widely. There are fusiform shapes like the human biceps brachii, that produce large-amplitude motion. There are also fan shapes, such as the pectoralis major,  that yield larger forces, sphincter morphologies that contract, and layered muscle sheets, like the transverse abdominis (Fig.~1A), that compress or transfer forces around the torso \cite{cutter1852treatise}.  
	The great variety of biological muscles, and their integration with other tissues, can inspire the design of soft robotic actuators, but much of the huge potential design space remains unexplored.
	
	Here, we describe a new family of muscle-inspired actuators that we refer to as Fluidic Fabric Muscle Sheets (FFMS).  They are composite fabric sheets that employ an integrated fluidic transmission comprised of arrays of hollow elastic tubes to generate  in-plane stresses or strains. We show how to design and fabricate these devices using apparel engineering methods. As we demonstrate, FFMS are stretchable, conformable, safe, efficient, and scalable. In order to situate this work relative to prior research, we begin with a review of several related technologies.

	\subsection{Background}
	
	Many soft actuator technologies have been developed for applications in robotics, wearable devices, healthcare, and other areas.  The FFMS sheet actuators we present here build on prior research on soft fluidic actuators, including the  sheet-like actuators described below.  It is useful to compare the materials, operating principle, methods of design and fabrication, and performance characteristics of such devices.  We highlight several salient examples below.

	\subsubsection{Fluidic actuators}
	
	Fluidic actuation technologies have attracted considerable attention for use in soft robotic systems because of their intrinsic compliance and the ease with which the fluids that transmit stresses may be integrated into soft media \cite{marchese2015recipe}. Fluidic power may be delivered in the form of fluid pressure and volume changes via a variety of pumps or charged reservoirs,  enabling  such systems to be designed to match a wide range of requirements.  Such devices can produce larger forces, displacements, or work densities  than are feasible with many emerging functional materials technologies (discussed below), facilitating practical applications. 
	
	Pneumatic Artificial Muscles, which generally shorten when filled with compressed air, come in many shapes and sizes \cite{daerden2002pneumatic}. Among the many soft fluidic actuators described in the literature, an early, influential example is the McKibben actuator \cite{schultecharacteristics, gavrilovic1969positional,chou1996measurement,tondu2000modeling}. It is a  soft, pneumatic device that comprises an airtight bladder with fiber constraints that cause it to contract when the internal volume is increased through the application of pressure.  The dimensions and working pressures of such actuators may be selected to match application performance requirements. However, maximum strains are typically limited to less than 35\% \cite{tondu1995theorie,hunter1991comparative}, limiting applications. Many soft, fluidic actuators  have been designed for pneumatic operation via compressible gases. This causes energy to be stored in gas compression during operation, which can lead to undesirable, rapid energy release on failure. Gas compression also leads to thermodynamic losses. A smaller group of soft, fluidic actuators have employed incompressible fluids. FFMS actuators can use either approach. We review some of the advantages of hydraulic operation below.

	Many other variations on the idea of combining fluidic actuation with fiber constraints have been investigated \cite{bishop2012design, connolly2015mechanical}. Recently, it has been observed that similar methods can be used to realize much larger strains -- approaching 300\% -- by means of actuators that contract when the internal volume and pressure are reduced, in a manner inverse to the McKibben design. Such devices, including the Inverse Pneumatic Artificial Muscle (IPAM) \cite{hawkes2016design} and Hydro-Muscle Actuators \cite{sridar2016hydro}, integrate anisotropic components that cause them to lengthen when pressurized. %, similar to the anisotropic behavior of certain plant cells that lengthen without swelling when pressurized. 
	We leverage just such an ``inverse'' fluidic actuation strategy in the FFMS actuators described here.  In contrast to the typically uniaxial and tubular forms of prior devices, which evoke fusiform muscle structures, FFMS actuators are actuated fabric-based structures, similar to muscle sheets.  Anisotropic constraints in FFMS actuators  are provided by the fabric structure.  Different fabric patterns and assemblies can be used to realize a variety of actuation modes, including uniaxial actuation, bending, multi-axis actuation, shape-changing, and compression. 
	
	Several methods for creating mechanical anisotropies with fabrics have been previously investigated to improve the performance of fluidic actuators, such as devices based on individual tubes \cite{sridar2016hydro}, air bladders \cite{cappello2018exploiting}, or other structures. In most cases, this is achieved through the intrinsic anisotropy of integral fibers. Fiber-reinforcement of the elastomer can be designed to produce desired anisotropy, and hence motion, by specifying the threading angle of the fibers \cite{bishop2012design, connolly2015mechanical}, although this complicates fabrication.  
	
		Other soft, fluidic actuator designs, including  origami-inspired devices, grow longer when positive pressure is applied \cite{martinez2012elastomeric}.  However, the maximum forces that can be reproduced are limited by buckling instabilities \cite{li2017fluid}. 	Vacuum-driven soft actuators have also been realized, attaining large peak stresses \cite{yang2016buckling}, but often involve large changes in cross-section area.

	\subsubsection{Other Transduction Principles for Soft Actuators}
	
	Many other methods of soft actuation have been investigated, each involving different tradeoffs in performance.  Shape memory alloys yield strains of up to about 5\% when heated \cite{otsuka1999shape}. They can also yield larger strains in other configurations, such as coils \cite{kim2009micro}.  Other thermally actuated transducers have been based on shape memory polymers, nylon, polyethylene, or other fibers \cite{haines2014artificial, li1998shape, takashima2010mckibben}. Such actuators often yield low efficiencies or low actuation speeds due to the intrinsic physics of the thermal processes underlying actuation. Other devices based on shape memory polymers \cite{liu2007review,zhang2007bending} or electroactive polymer technologies, including ionic polymer-metal composites \cite{jung2003investigations,guo2006underwater,bhat2004precision} have been designed to  yield high strains, but typically only generate small forces. Soft, electrostatic actuators, including dielectric elastomer actuators, are fast and can be designed to produce large strains \cite{carpi2011dielectric} but require high voltages and carefully controlled fabrication processes and mechanisms that preclude their use in some applications.  Variations on such actuators that use fluidic electrodes overcome some of the fabrication and design challenges involved in employing such actuators, but high voltages are nonetheless required \cite{kellaris2018peano}. Electromagnetic soft actuators can operate at low voltages \cite{guo2018liquid,do2018miniature}, but often require an external magnetic power source \cite{pawashe2009modeling, dd123development}.

	A more conventional approach to producing high forces and strains in soft, actuated structures is based on tendon- or Bowden-cable transmissions \cite{hofmann2018design,in2015exo,wei2018design}. Achieving high performance actuation and control with such devices depends on cable routing and friction management.  In addition, careful design is needed to ensure that the stresses that are produced are appropriately distributed.

	\subsubsection{Sheet-Like Soft Actuators}
	
		Various sheet-like soft actuators have been developed using  actuation methods paralleling those described above. Several groups have produced such sheet-like actuators based on shape memory alloys or thermally actuated nylons, but the tradeoffs between actuation time, forces, and strains limit the feasible mechanical power, speed, and reduce efficiencies \cite{yuen2016active, masuda2004preliminary, kobayashi1992woven}. In addition, the temperatures or heat exchange requirements may limit wearable applications.

	Multi-layered artificial muscles made of electrostatic sheet actuators have been designed to produce large forces at moderately high voltages, but require careful control over their motion during actuation, precluding out-of-plane deformation \cite{niino1994electrostatic}. Dielectric elastomer actuation principles have also been used to realize compact sheet-like soft actuators, although large voltages (greater than 5 kV) are normally required \cite{shian2015dielectric}.
	
	Sheet-like soft actuators have also been designed using fluidic actuation principles. This is achieved by assembling discrete pneumatic artificial muscles within a fabric or other sheet-like assembly, or through the design of integral fluid-driven cavities. One configuration comprised assemblies of McKibben muscles within fabric layers \cite{booth2018omniskins}. Another consisted of  thin McKibben muscles that were woven into fabric structures \cite{funabora2018flexible}. When driven to produce contraction along the axis of each muscle, the actuators in such devices expand, causing undesired increases in thickness, adversely affecting potential conformable or wearable applications and reducing efficiency.  Other designs have resulted in low forces that preclude many applications \cite{watanabe2018suitable}. In contrast, the FFMS actuators presented here increase in nominal length, or decrease in contraction force, when fluid pressure is applied.  This is achieved with negligible tangential expansion over the normal operating range of the actuator. 
	
	Other authors have used parallel cables or strings routed within fabric structures in order to achieve composite, sheet-like actuators \cite{shah2019morphing}.  Such designs can produce thin, fabric-like actuators, but require careful attention to cable actuation and friction management in order to ensure that dynamic, fast, and reversible actuation is possible.  Because FFMS actuators transmit stresses via integral fluids, losses, due to viscosity and channel length (see Modeling, below) remain within ranges that permit highly dynamic operation.

	\subsection{Contributions}
		
This paper presents FFMS, a new soft actuation technology. FFMS are planar, multimodal soft actuators that are analogous to muscle sheets.  Their design also builds upon prior ``inverse-type’' uniaxial actuators with tubular shapes that can be compared to fusiform muscles.  Here, we describe the design and fabrication of FFMS and show that this planar paradigm opens many new actuation capabilities and applications.

\paragraph*{Design, Fabrication, and Modeling:}
We show how to design composite fabric structures to realize mechanical anisotropies that enable FFMS to generate patterns of local contractions in a conformable, planar surface as fluid is withdrawn. FFMS can be efficiently fabricated using apparel engineering methods including pattern making, computerized sewing, and wrinkling, and through the integration of a fluidic transmission based on hollow elastic tubes. We describe several alternatives for their design and assembly.  We analyze the effects of fabric selection and wrinkling, tube routing, thread selection, and stitching selection, all of which can be used to tailor functionality and performance. We also present a simple mathematical model that proves effective for predicting their performance and aiding design. 

\paragraph*{Actuation Capabilities:}
FFMS actuators may be scaled to different sizes, depending on design requirements. We demonstrate actuators with dimensions ranging from 1 to 34 millimeters in thickness, and 30 to 1000 millimeters in length, yielding forces that can exceed 150 Newtons, and that can produce forces more than 115 times greater than their weight. FFMS actuators can also produce uniaxial engineering strains exceeding 100\%. Laboratory prototypes perform consistently in durability testing during 5000 cycles or more, with less than 5\% variation in displacement.

\paragraph*{Applications:}
We demonstrate their use in actuating  various bodies and mechanisms, and in  configurations that perform multi-axis actuation, including in- or out-of-plane bending and shape change.  We also demonstrate applications of FFMS methods for realizing low-profile, fabric-based actuators for new devices that exert forces on body tissues for wearable and biomedical applications. These include a glove for grasp assistance, devices for compressing small body regions or tissues, and devices for providing haptic compression or skin stretch to a finger, arm, or leg.  Compression garments formed from these actuators can produce dynamic compressive pressures that easily exceed 4000 Pa, which is sufficient for haptic feedback. Such compressions also meet requirements needed for many musculoskeletal, circulatory, wound, and lymphatic compression therapies, including peristaltic compression modes.
		
	\begin{figure*}[t]
		\centering
		\includegraphics[width=170mm]{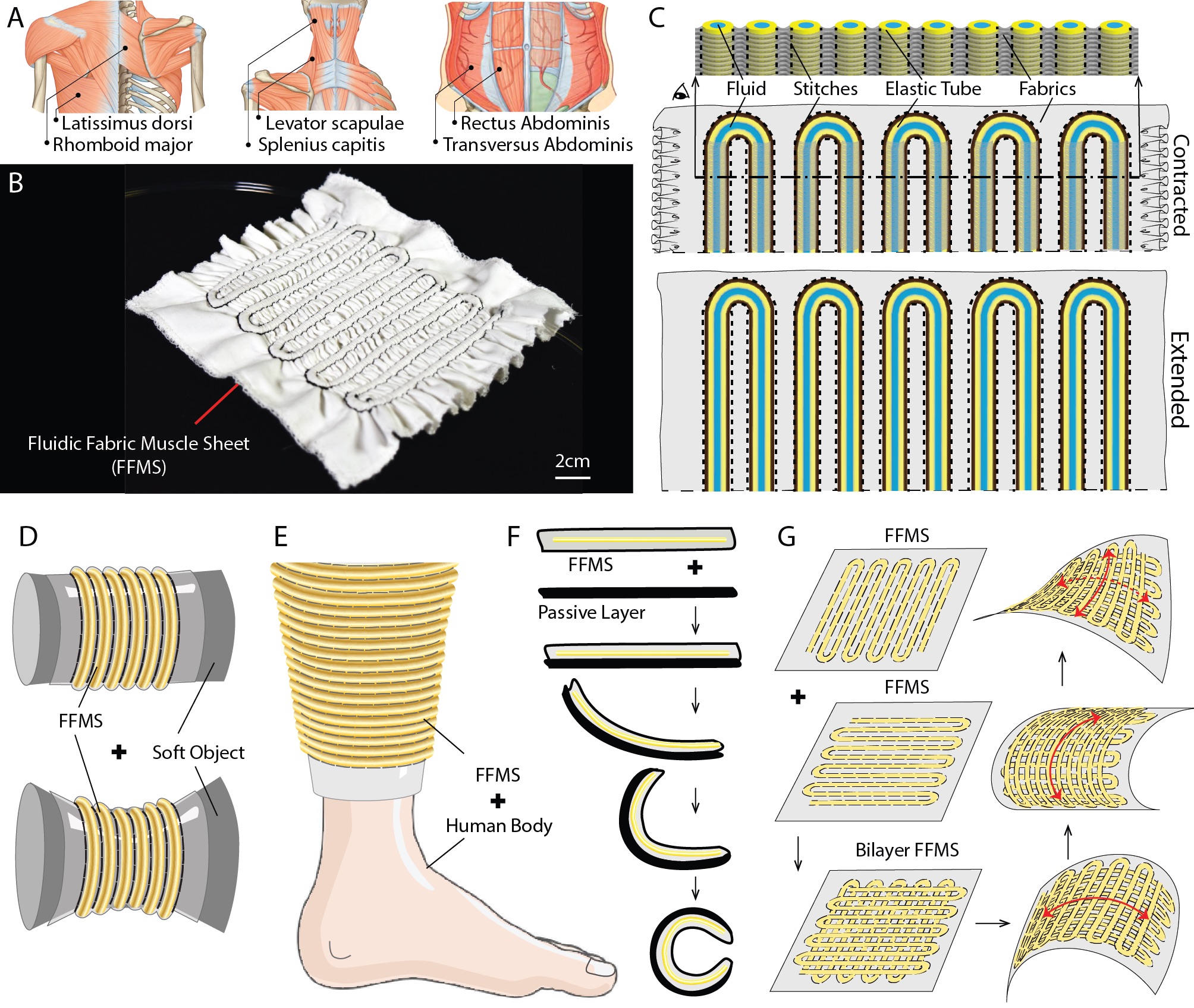} 
		\caption{
			A. Examples of muscle sheets in the human body that inspired the design of FFMS \cite{drake2014gray} (reproduction rights pending). B. A functional prototype illustrating how FFMS are planar fabric structures analogous to muscle sheets. C. FFMS comprise arrays of elastic tubes that function as fluidic transmissions. In this example, corresponding to the prototype of Fig.~1B, uniaxial extension is produced when fluid pressure is increased. D-G. FFMS may be applied in a variety of ways (see Fig.~\ref{fig:application}): (D) deforming a soft object, (E) compressing a limb, (F) bending a flexible structure, (G) in bilayer structures that generate morphological change, among many other possibilities.
		}
		\label{fig:front} 
	\end{figure*}

	\section{Design Concept and Operating Principle} 
	\label{sec:concept}
	\begin{figure*}[t]
		\centering
		\includegraphics[width=170mm]{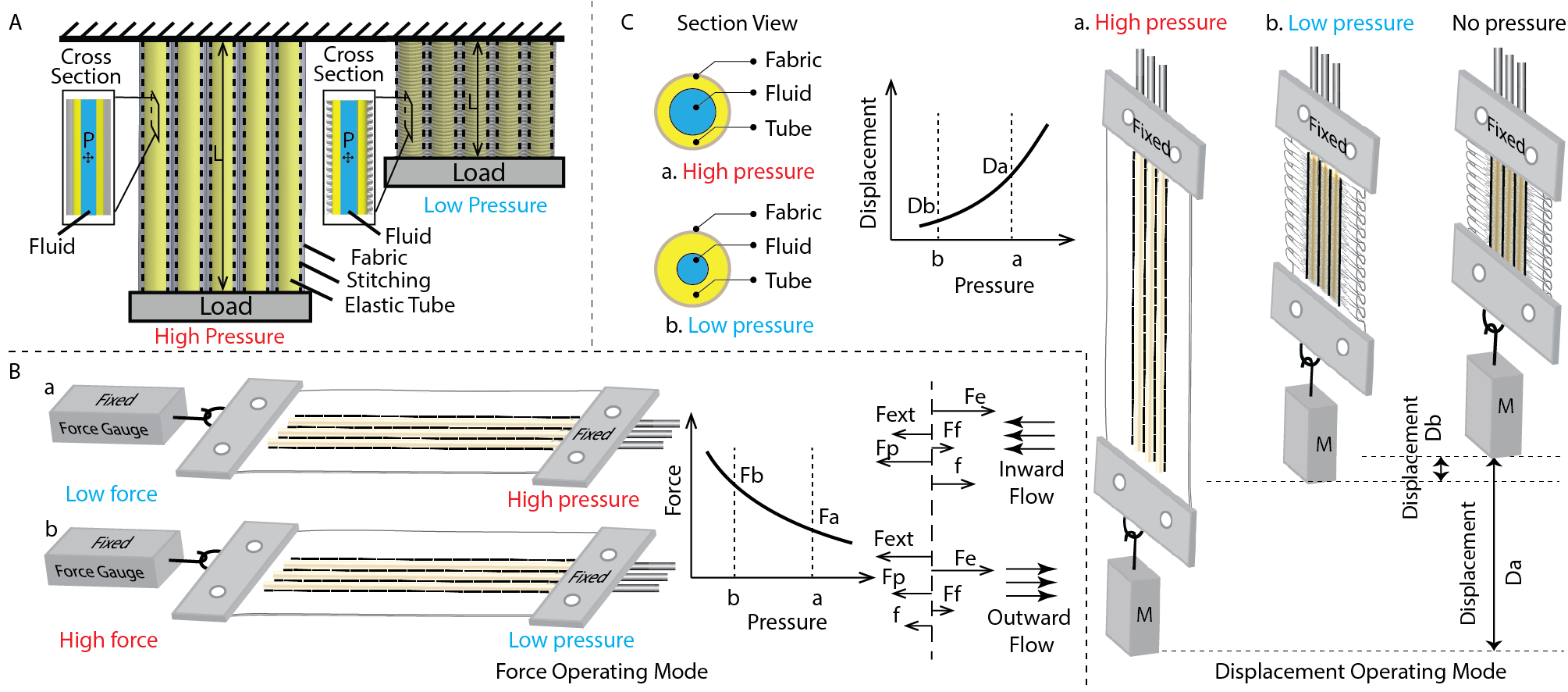} 
		\caption{
			Fluidic Fabric Muscle Sheets: Concept and operating principle. A. Hollow elastic tubes are integrated in a composite fabric structure. The tubes are routed in fabric conduits that provide circumferential constraints, due to stitching. \textit{Left:} When pressurized fluid is pumped into the elastic tubes; the tubes cannot swell radially due to the constraining stitches, and thus can produce a lengthening, similar to a relaxing sheet of muscle. \textit{Right:} When the fluid pressure is removed, stored elastic energy in the hollow tubes and elastic fabric is released, and the entire textile shortens, like a contracting sheet of muscle. $P$ and $L$ represent fluid pressure and actuator length, respectively. B. \textit{Left:} When the FFMS is operated to work against a load, as in the isometric configuration shown here, forces are produced.  \textit{Middle:} High pressures produce low forces, and vice versa. \textit{Right:} A simple illustration of the generation of axial forces.  We present a mathematical model in a subsequent section. C. \textit{Left:} Cross section view of a single channel in the displacement operating mode. \textit{Middle:} In displacement mode, higher pressures produce larger displacements, and vice versa. \textit{Right:} In such a displacement mode, the FFMS may be used to do external work, such as lifting a mass, as shown here.
		}
		\label{fig:working_principle} 
	\end{figure*}

	The FFMS actuators presented here generate stresses or strains in a composite fabric sheet when charged with a pressurized fluid.  The fluid may be compressible (pneumatic operation) or incompressible (hydraulic operation); we compare the relative merits of each approach later in the paper.  The routing of stresses and strains is accomplished via an integral fluidic transmission composed of hollow elastic tubes.  The fabric assembly imposes mechanical anisotropies that locally direct stresses along axis of channels that are sewn into the fabrics.  As fluid is pumped into the elastic tubes, their internal volume is forced to increase. Circumferential constraints imposed by the fabric cause the increased fluid volume to produce a nominal lengthening of the structure along the axis of each tube (Fig. \ref{fig:working_principle}A). This causes elastic energy to be stored in the tube-fabric structure. As the fluid pressure or volume is reduced, the stored elastic energy is released. When the FFMS is working against a load, this reduction in pressure yields a contraction force.  Thus, while increased pressure produces a lengthening of the FFMS, external forces are normally produced via contraction, similar to biological muscles (Fig.~\ref{fig:front}A) \cite{drake2014gray}. As we demonstrate, in other configurations the same principle can be used to realize FFMS actuators that operate in hydrostatic mode, similar to muscular hydrostats such as the tongue of many animals or trunk of the elephant. The ranges of forces and displacements that can be produced depend on the dimensions of the FFMS,  operating range of applied fluids,  actuation mode, and materials involved.  We present a simple mathematical description of the effects of these factors below.
	
	Depending on actuation requirements, the FFMS can be operated to produce forces or displacements.   In force mode (Fig.~\ref{fig:working_principle}B), a muscle is pre-stretched against a load via constraints at both ends.  At high fluid pressures, the forces produced are low, while low pressures generate high forces. Stresses are produced along the longitudinal axis of the elastic tubes. This can yield axial forces at the ends of a fusiform-shaped FFMS, or compression forces, when the FFMS is wrapped in a nose-to-tail configuration around an object, such as a human limb (Fig. \ref{fig:application}F). FFMS may also be used to generate large strains or displacements (Fig. \ref{fig:working_principle}C). Changes in displacement can be used to displace loads or to alter the shape of a structure through differential stresses or strains.  Due to the radial constraint imposed by the composite fabric structure, a change in fluid volume creates a change in length. The relationship between the fluid volume in the FFMS and the channel length is approximately linear (see Section \ref{results}).  Such a displacement can be used to perform mechanical work.  As we show, in multi-layer structures, it can also be used to effect changes in the intrinsic shape of an FFMS assembly. Together, these unique capabilities make FFMS amenable to various applications.

	\section{Fabrication}
	
	\begin{figure*}[t]
		\centering
		\includegraphics[width=170mm]{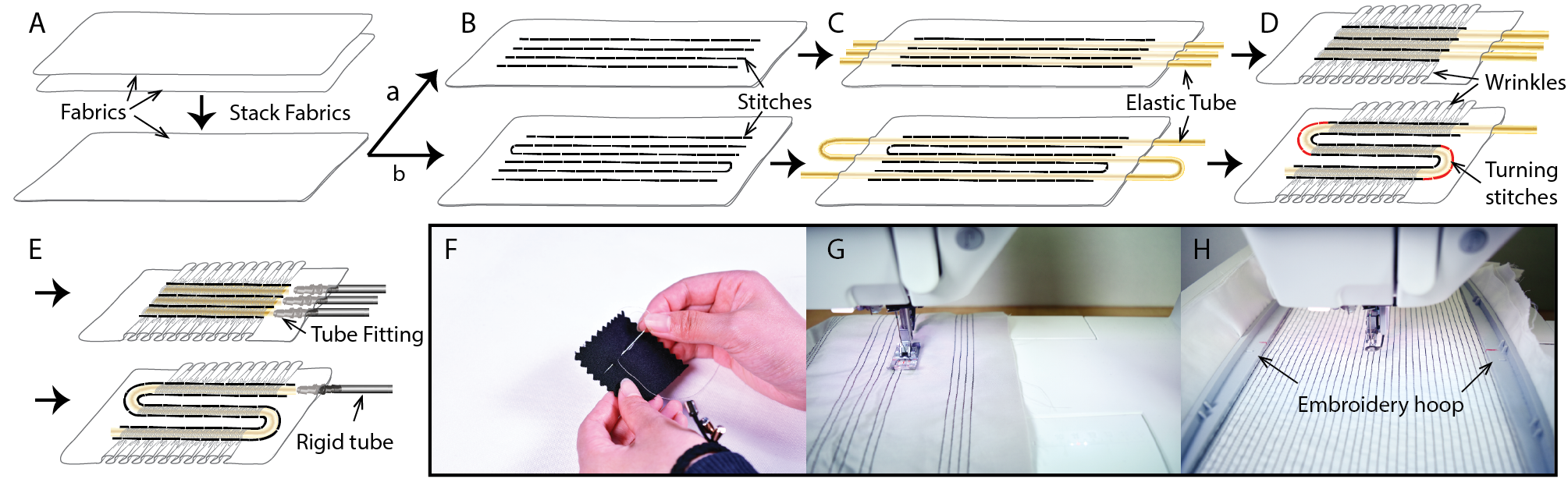} 
		\caption{
			Fabricating  planar fluidic fabric muscles involves several steps based, in part, on  apparel engineering methods. For configurations based on non-stretchable fabrics, the fabric layers are first aligned and stacked. (A) The routing of elastic tubes is designed and layers are stitched to form conduits   (B)  in a pattern that determines the routing. The stitched patterns can realize  single (b) or multiple (a) tube routings. The elastic tubes are then threaded (C) through the resulting fabric conduits. (D) For non-stretchable fabric layers, the fabric structure is  wrinkled. (E) A port is established at one end of each channel, whose remote end  is then sealed. The stitching may be done via (F) hand sewing, (G) machine sewing, or (H) computerized embroidery. 
		}
		\label{fig:fabrication_process} 
	\end{figure*}

	The fabrication of FFMS actuators involves three main steps: Construction, patterning, and assembly of a multi-layer textile structure, routing of elastic tubes in the patterned fabric structure, and sealing and attachment of tube fittings, as illustrated in Figure \ref{fig:fabrication_process}A-E. The figure highlights a process based on a configuration of non-stretchable fabrics with stitching parallel to the channels.  Other stitching patterns, which are amenable to designs  based on stretchable fabrics, are discussed in Section \ref{materials} below. 

	In a first step, the fabric layers are aligned and stacked. The layers are then stitched to form conduits in between the fabric layers that allow the insertion of elastic tubes. The stitching patterns can be designed to realize configurations using a single tube with a single fluid input port (Fig.~\ref{fig:fabrication_process}a), or multiple tubes with separate ports that allow the independent control of multiple channels (Fig.~\ref{fig:fabrication_process}b). The routing of channels in the fabric determines the distribution of strain and stress within the composite textile, which need not be uniform.  To radially constrain the tubes effectively, the conduit determined by the stitching is designed to have a width equal to  half of the tube diameter.  
	
	Different sewing methods can be used: hand sewing (Fig. \ref{fig:fabrication_process}F), machine sewing (Fig. \ref{fig:fabrication_process}G), and computerized embroidery (Fig. \ref{fig:fabrication_process}H). When the size of the pattern fits within the maximum embroidery hoop size that the sewing machine can accommodate, computerized embroidery  is preferred. It provides great accuracy, flexibility, and efficiency. When embroidery is impossible, machine sewing may be used. This involves manual movement of the fabric under the sewing foot. This is often the best option when stitching long actuators or very large surface areas.  Hand sewing is the least efficient method, but can accommodate complex paths or non-flat fabric surfaces, as are involved when stitching the turning stitches that we apply following tube insertion  (Fig. \ref{fig:fabrication_process}D). 
	
	After the sewing step, the elastic tubes are threaded through the stitched fabric conduits.  To achieve this, we use a slender rod that is inserted securely into one end of the elastic tube. Once the tubes are completely inserted, the fabric is then wrinkled along the length of the tubes in order to accommodate stretching in that direction. We have found that this wrinkling can readily be performed in a uniform, controlled fashion.  Together, the proportion of wrinkling and elastic properties of the fabric and tube determine the maximum stretchability of the muscle. This process is best suited to non-stretchable fabrics. If stretchable fabrics are used, the wrinkling step may be omitted. In a next step, one end of the elastic tube is sealed with a fixture, such as a knot,  while the other is connected to a barbed tube fitting that allows the fluid to be supplied to the muscle. For  large-scale  FFMS (Fig.\ \ref{fig:scalability}A,B), one end of the tube may be sealed with a solid barbed end plug, rather than a knot, while the other end is connected to a barbed tube fitting. To strengthen the connection between the fitting and the elastic tube, clamps may be applied.  In a next step, air is purged from the channels. To stabilize the mechanical response, the muscle should be fully extended and contracted several times prior to the first usage.

	%	After that, one end of the elastic tube was sealed through the application of a tight knot or a tube fitting cap. 
	
	\section{Material Selection}
	\label{materials}
	The performance of FFMS actuators depends greatly on the selection of elastic material, fabric material, and stitching pattern. The working ranges of displacements and forces are determined by the applied fluid pressure range, the fabric and elastic tube material properties and sizes, and the manner of patterning and assembly.

	\subsection{Fabric and Stitch Selection} 
	
	Forces and motions produced by the FFMS are determined by the relative magnitude of stored energy generated by the elastic tubes and the fluid pressure. To achieve the desired dynamic range of forces and motions, the axial stretchability of the fabric conduit should be maximized, while the radial expansion of the elastic tube in the operating range of fluid pressures should be minimized. A firm circumferential constraint is needed to ensure that stresses due to the fluid are directed along the axis of the tube and do not result in the tube expansion. Based on these criteria, an ideal fabric structure should possess negligible stiffness in the axial direction of tube so that a high elongation from the muscle can be achieved (we provide a quantitative discussion in the modeling section, below). This requirement can be achieved through the use of non-stretchable fabrics, such as cotton weaves,  in tandem with the wrinkling process that is applied during assembly. In some applications, stretchable fabric may be of interest.  In such cases, cross stitching may be used to impose a radial constraint. Such stretchable fabrics are uniaxially elastic (two-way stretch) or biaxially elastic (four-way stretch).  These fabrics are often made of elastic fibers such as Spandex, that are spun into stretchable yarn, and integrated along weft, warp, or both directions of the weave, yielding one- or two-way stretch fabric, respectively. Alternatively, either elastic or non-elastic fibers may be used to create a knit. In a knit, stretchability depends on the design of the looping structure. This typically results in biaxial stretchability. The axial stretchability of FFMS fabric structures using stretchable fabrics can also be improved through the application of wrinkling, although we have not encountered practical situations for which this is needed.
		
	The routing of stresses or strains within the FFMS is achieved via fabric conduits formed from stitching applied to the fabric sheets. The stitching involves three main factors: thread material, stitch type, and stitch pattern. Near the elastic tubes, the stitching permits the fabric to impose a circumferential constraint. This ensures that fluidic stress or strain is directed along the axis of the tube  (see Section \ref{sec:concept}). Two design criteria are involved.  First, the stitching must be strong enough to constrain the fabric conduit around the elastic tubes over the entire operating regime of the FFMS. Second, it must accommodate large strains along the axial direction of the tube, even in the absence of wrinkled structure. Several different stitch designs, together with different choices of fabric, can meet these requirements.  The selection of each depends on  application requirements. To illustrate this, we compare three different combinations of fabric type and stitch design in Table \ref{tab:StitchandFabrics}. This table shows that combinations yield distinct patterns of stretchability in the assembled FFMS (Table \ref{tab:StitchandFabrics}, blue arrows).

	If a wrinkling step is omitted, a two-way or four-way stretchable fabric must be used in order to accommodate the axial strains that are required for a FFMS actuator to function.  A side or cross stitch pattern may be used with either stretchable or non-stretchable fabrics, with or without wrinkling constraints (Table \ref{tab:StitchandFabrics}, red arrows).  Cross stitching is preferred for use with four-way stretch fabrics, because it  minimizes undesired radial expansion (i.e.\ ballooning) of the elastic tubes. Significant ballooning only occurs if side stitch patterns are used. Commercial two-way stretch fabrics (which are typically knits) admit fabric extension in all directions.  Thus, such fabrics yield undesirable ballooning unless a cross stitch is used, which can result in failure (see Fig. \ref{fig:failure}C).  When non-stretchable fabric is used, wrinkling must be applied. In such cases, the range of extension is determined by the extent of wrinkling.  In practice, we have designed FFMS actuators capable of greater than 300\% strain using this technique.  If a cross stitch is used, the amount of wrinkling is limited due to the increased fabric constraint imposed by the cross stitch, thus limiting the extensibility of the FFMS.

\begin{figure*}[t]
		\centering
		\includegraphics[width=83mm]{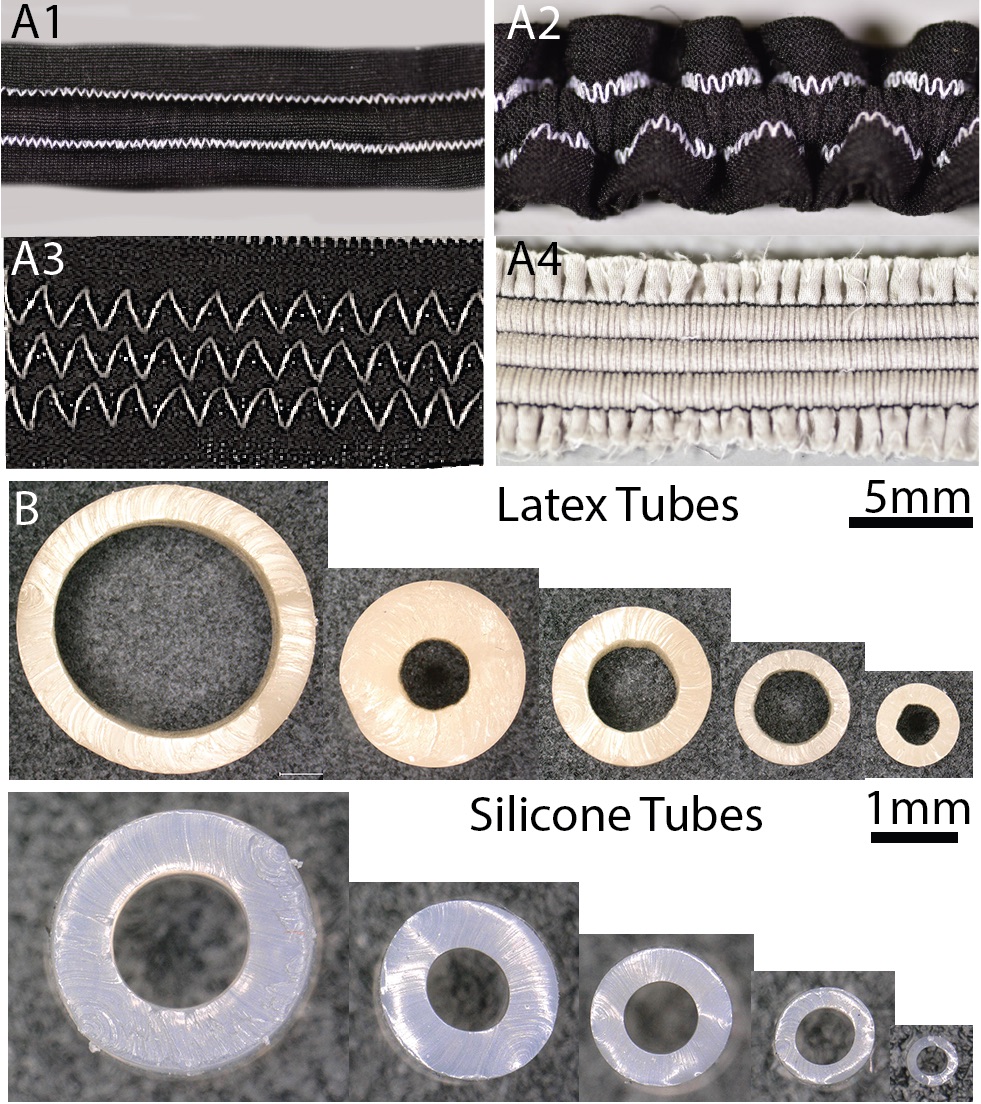} 
		\caption{
			A1-A4. Four representative FFMS prototypes illustrating different fabric and stitch combinations. A1. Two-way stretch fabric using zig-zag side stitches without wrinkling. A2. Two-way stretch fabric using zig-zag side stitches with wrinkling. A3. Two-way stretch fabric using zig-zag cross stitches without wrinkling. A4. Non-stretch fabric using straight side stitches with wrinkling. B. Cross section images of several commercially available tubes made of latex (top) and silicone (bottom, note the smaller scale). The outer diameter of the silicone tube can reach sub-millimeter scales.
		}
		\label{fig:prototypesandtube} 
	\end{figure*}

Among thread types, inextensible high-strength nylon thread is one choice that is able to provide a sufficiently stiff constraint via a fine thread. There are several possible combinations of stitch designs, fabric stretchability, and wrinkling modes, which together yield distinct patterns of stretchability in the assembled FFMS (Table \ref{tab:StitchandFabrics}, blue arrows; longer arrows imply larger stretchability).  When side stitches are used, zig-zag stitching is recommended for use with stretchable fabrics in order to preserve the fabric elasticity.  Straight stitching is appropriate for non-stretchable fabrics, where stretching is accommodated by wrinkling. To maximize stretchability in the axial direction, and minimize radial expansion, two-way stretch fabric with side stitches and wrinkling is optimal, although non-stretch fabric with side stitches and wrinkling is also effective. %For example, a flatter or thinner FFMS can be achieved if no wrinkling is applied. In this case, stretchable fabrics may be the best option. 
	Figure \ref{fig:prototypesandtube}A shows four prototypes with different combinations of fabrics and stitchings.

	\begin{table}[t]
		\centering
		\includegraphics[width=170mm]{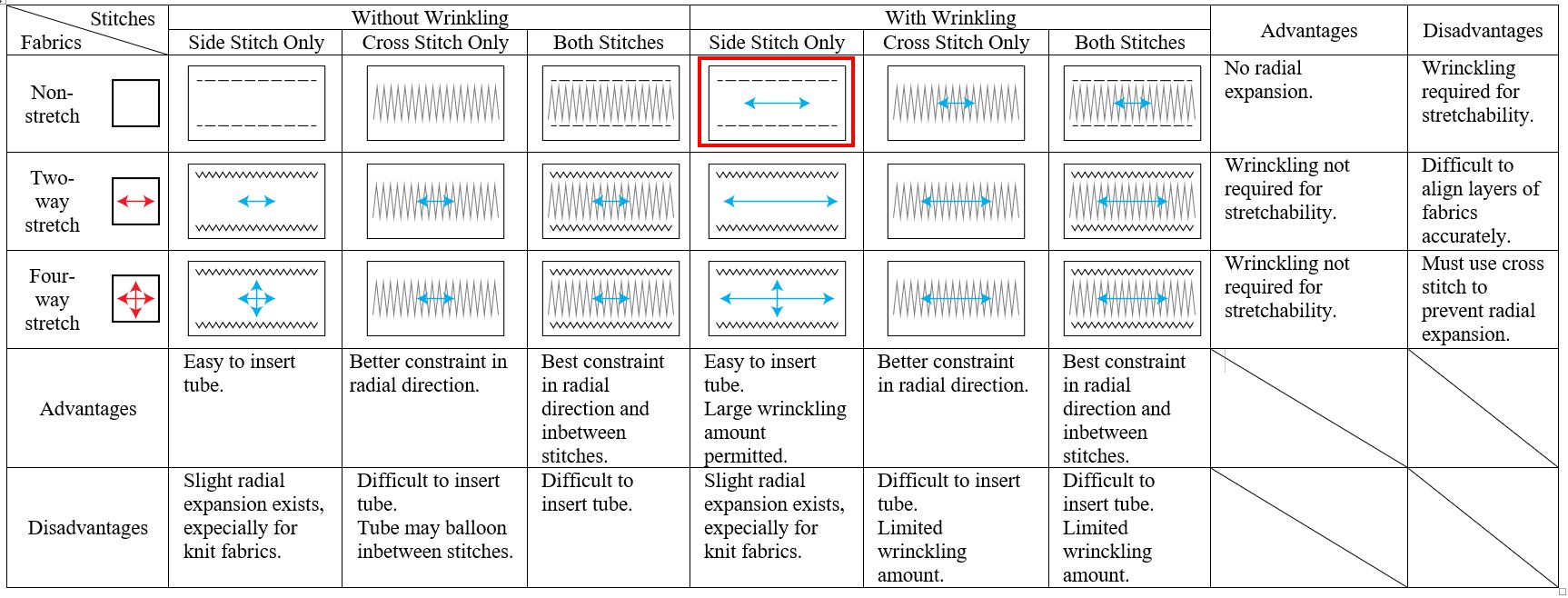} 
		\caption{
			Stitch and fabric selection for FFMS actuators. Red arrows represent the stretch directions of the fabrics, and blue arrows represent the stretchability of the assembled FFMS. Longer arrows denote greater stretchability. The configuration in the red box is used for most prototypes in this work.
		}
		\label{tab:StitchandFabrics} 
	\end{table}

	\subsection {Tube and Fluid Selection}
	Another important material to specify is that of the elastic tube.  This may be any elastomer that is compatible with the working fluid. For high force applications, tubing materials with high elastic modulus and high extensibility would be preferred. The tube can have any diameter and length compatible with the fabrication process, including very fine silicone tubing \cite{do2017stretchable}, or larger diameter latex rubber tubing.  As noted below, FFMS using larger tubes generate higher force per unit length on the threads, which can lead to failure for very large tubes. 
	There are many options among commercially available tubes (Fig.\   \ref{fig:prototypesandtube}B). We present a model of the actuator performance accounting for  effects of tube size and material properties in Section \ref{model} and present  experimental results for several examples in Section \ref{results}. %A force prediction in relation to the tube size and material properties is further discussed in the modeling section.

A variety of working fluids may be used, depending on the performance requirements, materials, and operating criteria. In hydraulic operation, FFMS can use incompressible fluids such as oils or waters, as illustrated in many of our experiments. This makes it possible to control the applied volume in the actuator and hence the actuator length, due to the circumferential constraints. This renders the quasi-static response of the actuator very simple, at the expense of increased viscosity and mass (however, the fluid mass in many of our prototypes is on the order of a few grams). Low viscosity fluids may be used to improve the actuation speed and efficiency. Pressures used with our prototypes are generally much lower (less than 0.75 MPa) than those used in industrial hydraulics (20 MPa would be typical). At such low working pressures, failures typically involve, at most, fluid leakage. 

In pneumatic operation, compressible fluids such as air or other gases may be used. This can minimize  mass and viscosity.  Over the operating range tested in our experiments, pneumatic operation leads to increased hysteresis, lower efficiency, and increased response latency (Fig.\  \ref{fig:testing_Displacement}F). % this can lead to increased hysteresis and may also introduce response latency. %Many pneumatic power sources and control systems operating at the pressures of interest are commercially available, the use of such compressible fluids can facilitate implementations of closed-loop control systems. 
Further discussion of hydraulic and pneumatic actuation methods are provided in the literature \cite{de2010pneumatic,polygerinos2017soft}.
	
	\section{Analytical Modeling}
	\label{model}
%	\subsection{Axial Forces}

As our experiments demonstrate, FFMS actuators may be operated to yield a variety of motion or force patterns.  The simplest involves the generation of axial forces through a parallel configuration of $N$ elastic tubes.  Such a structure is  similar to parallel muscle sheet architectures in biology.  A net external force, $F_{ext}$, is produced by the actuator due to the stretching of the tubes, which produce a net elastic force, $F_{el}$. The fabric can also contribute an elastic force,  $F_{fab}$. The force $F_{ext}$ exerted by the actuator decreases with increasing fluid pressure, $p$, due to the axial force, $F_{fluid}$, generated via the fluid pressure. Dissipative forces, $F_d$, include viscosity and friction.  Combining these factors, one can model the forces produced by the actuator as
	\begin{equation}
	F_{ext} = F_{el}+F_{fab}-F_{fluid} + F_{d} \ \ %\text{where} \  F_{d} = |F_{c}| \text{sgn}(v) \, .
	\end{equation}
The dissipative forces, $F_{d}$, include hydrodynamic flow resistance, $F_{d,hyd}$, and  dry friction at the tube fabric interface, $F_{d,dry}$
    \begin{equation}
    F_{d} = F_{d,hyd} + F_{d,dry}	
    \end{equation}
$F_{d,hyd}$ can be estimated from the Newton's Law of viscosity \cite{batchelor1967introduction}, 
    \begin{equation}
		F_{d,hyd} = \tau \mathcal{A}= \mu \rho \frac{ \partial u}{\partial y} 2\pi r_i L
    \end{equation}
where $\tau$ is the shear stress of the fluid acting on the inner wall of the tube, $\mathcal{A}$ is the contact area between the fluid and the tube, $\mu$ is the kinematic viscosity, $\rho$ is the fluid density, $u$ is the flow velocity, $\partial u / \partial y$ is the rate of shear deformation, $r_i$ is the tube inner radius,  and $L$ is the tube length. The dry friction $F_{d,dry}$ is given by
    \begin{equation}
	F_{d,dry} = \mathcal{F_N} \zeta  = N \zeta p \, A_{tf}, \ \ A_{tf} = 2 \pi r_o \,L = N \zeta p \, A_{tf},
    \end{equation}
 where $N$ is the number of tubes, $\zeta$ is the friction coefficient, $\mathcal{F_N}$ is the normal force between the tube and fabric, $A_{tf}=2\pi r_o L$ is the area of the tube-fabric interface, $r_o$ is the tube outer radius, and $p$ is the fluid pressure. We refer to such dissipative forces when interpreting measurements of actuator efficiency and hysteresis.

 In many embodiments, including the wrinkling construction described above, there is a small relative motion of the tube and fabric, so  $F_{d,dry}$ may be neglected.  In quasi-hydrostatic operation, the forces due to flow resistance $F_{d,hyd}$ may also be neglected.  For fabric that is wrinkled or easily stretched in the axial direction, the fabric force, $F_{fab}$ may also be neglected.\footnote{Selecting a stiffer fabric, or adding non-fluidic elastic fibers to the structure, increases forces, but does not necessarily improve actuator performance, because the added elasticity does not impart any added strength to the tube that would enable it to operate at  higher pressures.}  Assuming that these conditions hold, and assuming that the FFMS operates in a linear elastic regime with elastic modulus $E$ and true strain $\epsilon$, the net external force exerted by the actuator is:
 \begin{eqnarray}
 	F_{ext} &=& F_{el}-F_{fluid}	\, , 
 \end{eqnarray}
 where
  \begin{eqnarray}
	F_{el} \ \ \ &=& N  E \epsilon A_{tube} = N E \epsilon \, \pi (r_o^2 - r_i^2) \\\
 	F_{fluid} &=& N p A_{fluid} = N p \pi r_i^2 \, 
 \end{eqnarray}
The result may be written
%\begin{equation}
%\epsilon = \int_{L_0}^L \frac{dL'}{L_0} = \log\left(\frac{L_0 + \delta L}{L_0}\right)
%\end{equation}
%where $L_0$ is the rest length and $\delta L$ is the length change. 
  \begin{eqnarray}
 	F_{ext} &=& N  ( E \epsilon A_{tube}-  p A_{fluid})\,
 	  \label{eq:Fext}
 \end{eqnarray}
The external force $F_{ext},$ reaches its maximum value if the fluid pressure $p=0$, 
\begin{equation}
\max_p F_{ext} \ = \ F_{ext} \vert_{p=0} \ = \ N E \epsilon A_{tube}\,
\end{equation}
The force can be maximized by increasing the value of the net cross section area, $N A_{tube}$, the elastic modulus, $E$, or the operating range of strains, $\epsilon$. Increasing the strain may be accomplished via pre-tensioning, which is also aided by wrinkling. The model given by Eq.~\ref{eq:Fext} shows that $F_{ext}$ decreases linearly with the increase in pressure, $p$.  As our experiments demonstrate, despite the simplifications involved in this model, it provides a good fit to the behavior of FFMS actuators (Sec.\ref{results}, Fig.\ref{fig:testing_Force}D4 and G4).

The minimum external force magnitude occurs when the pressure is maximum, $p_{max}$. Since pressure fluctuates in dynamic operation, we take this to be the maximum quasi-hydrostatic pressure. If the operating range of forces extends down to $F_{ext}=0$, the required maximum pressure, $p_{max}$ is determined by the ratio of tube and fluid areas and tube elasticity, 
 \begin{eqnarray}
 p_{max}  &=& \frac{ E \epsilon A_{tube}}{A_{fluid}} \, . 
 \label{eq:pmax}
 \end{eqnarray}
where $A_{tube}$ is the inner cross section area of the tube. Conversely, if $p_{max}$ is the maximum intended pressure, the tube geometry, strain, and elasticity should be selected to ensure this expression holds.  

In another configuration, the actuator may be operated as a muscular  hydrostat \cite{kier1985tongues}.  For unsupported (isochoric) operation, or for negative contraction forces, $F_{ext} < 0$, the muscle force may be produced by applying a pressures higher than the one specified in  (\ref{eq:pmax}). In the absence of an external load force, $F_{ext}=0$,  a pressure $p$ applied to hydrostatic configuration can yield a displacement, $\delta L,$ with respect to the initial tube length, $L_0,$ 
\begin{equation}
\delta L = L_0 \exp\left(\frac{p}{E}\right) - L_0    
\end{equation}
 Such modes of operation are analogous to muscular hydrostats in biology, such as the tongue of many animals, or the elephant trunk.

FFMS actuators may also be used to compress enclosed objects.  If an actuator  with pressure $p>0$ perfectly encloses a rigid cylindrical object, compressive pressures are generated as the fluid pressure $p$ is reduced.  For a cylinder of radius $r_c \gg h$, where $h$ is the effective FFMS thickness, the compressive pressure, $p_c$, or force per unit area exerted on the cylinder, is
\begin{equation}
    p_c = \frac{hF_{ext}}{r_c  A_M}, 
\end{equation}
%where $\sigma = F_{ext} / A_{M}$, 
where $A_{M}$ is the net effective  cross section area of the FFMS.

	\section{Results}

To evaluate the proposed FFMS actuators, we performed mechanical testing in several experimental configurations and operating modes, using several FFMS actautors of different sizes.  We also realized functional prototypes for wearable devices, haptic feedback, and soft robotics.

	\label{results}
	\subsection{Mechanical Testing}

We used three testing configurations to measure axial forces, compressive forces, and axial displacements. We complemented these evaluations with measurements of  actuation efficiency and durability over thousands of cycles. 

Detailed mechanical testing was performed using two different prototypes (Fig.~\ref{fig:testing_Force}). The first comprised a smaller surface area FFMS with three elastic tubes (Fig. \ref{fig:testing_Force}B) of dimensions  196 mm (length), 25.2 mm (width), and 4.7 mm (thickness).  The fabric channel width was 5 mm.  The active area spanned by the elastic tubes was measured to be 122.4 mm in length with no applied pressure.  The tubes were connected via a manifold to the fluidic power source.  We used latex tubes with outer diameter 3.2 mm and inner diameter 1.6 mm.

The second prototype consisted of a larger surface area FFMS with ten parallel elastic tubes   (Fig. \ref{fig:testing_Force}C).  The tubes in this prototype were connected in series, yielding a single fluid port. The dimensions of this prototype are 148.4 mm (length), 156.6 mm (width), and 4.9 mm (thickness) while the channel width is 5 mm.  The active region spanned by the elastic tubes (neglecting  unwrinkled fabric end sections) was 84.1 mm.  

In all experiments, distilled water was used as a working fluid. In a separate experiment, we compared the operating efficiency of water (hydraulic mode) and air (pneumatic mode).

	\subsubsection{Axial Force Testing}
\begin{figure*}[t]
		\centering
		\includegraphics[width=168mm]{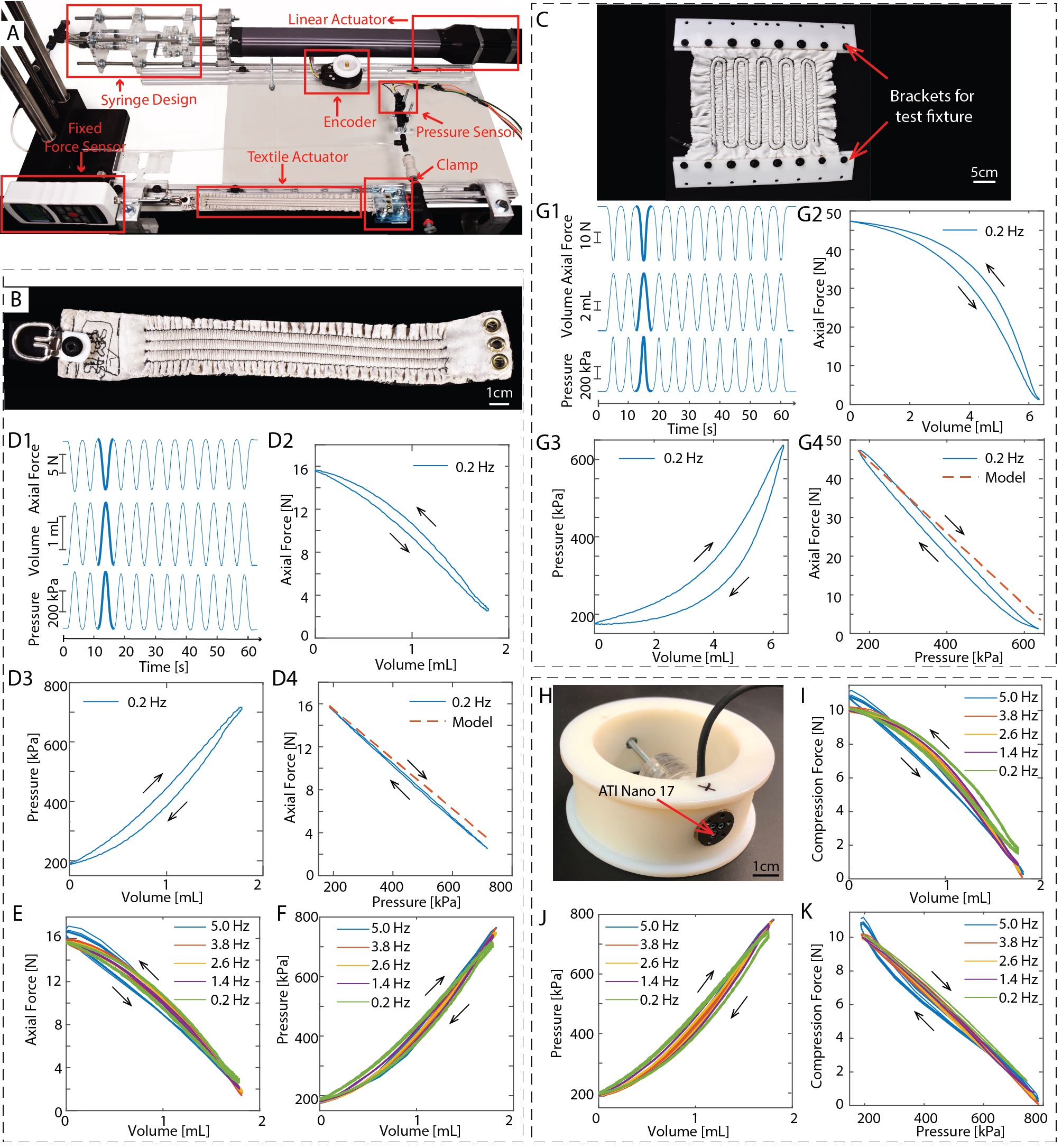}
		\vspace{-4mm}
		\caption{
			Force testing.   A. Apparatus for axial force testing. B. 3-channel FFMS. C. 10-channel FFMS with brackets for test fixture. D-F. Results for the 3-channel FFMS demonstrated consistent performance over repeated actuation, similar behavior at different actuation speeds, and surprisingly good agreement with the analytical model.  G1-G4. The larger, 10-channel FFMS yielded similar results to those that we obtained with the 3-channel device. The force range was 0 to 50 N. The longer fluid circuit yielded slightly greater response latency. H. Compression force testing apparatus. I-K. The device produced compressive forces of 0 to 10 N, as 0 to 2 mL of fluid was withdrawn.  The results were consistent with our analytical model, and varied little with speed.
		}
		\label{fig:testing_Force} 
	\end{figure*}

Axial force testing was performed using an isometric test configuration and apparatus (Fig.\ \ref{fig:testing_Force}A). One end of the FFMS was fixed using a clamp fixture while the other end was attached to a stationary force gauge (M5-20, Mark 10).  Fluid was supplied via three 10 mL syringes (inner diameter around 15 mm) driven by a displacement-controlled linear motor (A-BAR300BLC-E01, Zaber) and custom fixture. The syringe displacement was also measured via an optical encoder (S6S-1000-IB, US Digital). Fluid pressure was measured using a fluid sensor (SSC Series Sensor, Honeywell) positioned near the actuator port. The actuator was pre-pressurized to approximately 650 kPa and fixed in an isometric configuration with sufficient tension to ensure that the actuator remained during  testing.

The FFMS was driven via sinusoidal fluid displacement, at frequencies from 0.2  to 0.4 Hz.  This yielded time-varying (measured) pressures ranging from 200 kPa to 750 kPa. The signals were decoded synchronously using a computer-in-the-loop system (QPIDe, Quanser, Inc., with Simulink, The Mathworks, Inc.).  The instantaneous displacement of fluid volume was calculated from the syringe displacement and geometry. 
	
The FFMS performance was consistent over repeated cycles. The generated axial force decreased monotonically with the increase of fluid pressure or volume (Fig.\ \ref{fig:testing_Force}D-G).  The results for a single cycle  (Fig.\ \ref{fig:testing_Force}D2-D4) indicate that the range of forces $F_{ext}$ generated by the FFMS was 13 N (approximately 4.3 N per tube). This corresponded to a fluid volume range of 1.78 mL. The volume-force and the volume-pressure curves both exhibit hysteresis, indicating that energy was lost on each working cycle. We will discuss such losses in the efficiency measurements  below.

We compared the results with predictions of the analytical model (Eq. \ref{eq:Fext}). To compute these predictions, we used the geometric dimensions and measured the Young's modulus of the elastic tube.  Using tensile testing, we determined the Young's modulus to be $1.1$ MPa for true strains, $\epsilon$, between 0 and 1. Other parameters used for  model prediction were $A_{tube}=5.9 \times 10^{-6}$ m$^2$, $\epsilon = 0.8$, and $A_{fluid}=7.7 \times 10^{-6}$ m$^2$. The number of tubes was $N=3$ for the first and $N=10$ for the second. The experimental results and model predictions were in  close agreement during slow actuation (Fig.\ \ref{fig:testing_Force}D4). 

For different frequencies, ranging from 0.2 Hz to 5.0 Hz, the FFMS actuator responded in a qualitatively similar manner (Fig.\ \ref{fig:testing_Force}E,F).  The response became somewhat more complex at the highest frequencies. This can be explained by the inherent dynamics of fluid-elastomer-fabric systems.

We obtained results when testing the larger FFMS actuator with 10 parallel tubes  (Fig. \ref{fig:testing_Force}G1-G4). In this case, the FFMS produced forces ranging from 0 to 50 N with respect to the decrease of fluid volume (from 0 ml to 6 ml) or fluid pressure (from 620 to 180 kPa). The FFMS performance was also consistent over repeated testing cycles.  Each of the ten tubes produced about 5 N.  For the smaller prototype, analytical modeling yielded qualitatively good agreement with the experimental results (see Fig. \ref{fig:testing_Force}G4). The results revealed nonlinear hysteresis between the pressure and force.  This can be attributed to the fact that the tubes were connected in series, requiring the fluid to traverse a much longer path.  This indicates that series connections of elastic tubes, which simplify assembly, may introduce modest response latency.  From the experimental data, we estimated the latency to be approximately 100 ms at 0.2 Hz.  The volume-pressure and volume-force relationships also exhibited nonlinear hysteresis, indicating that energy was lost on each cycle.  For small changes in fluid volume, from 0 to 2 ml, the change in force with fluid pressure was more gradual.  However, the pressure-force relationship remained approximately linear over most of the testing range.

	\subsubsection{Compression Testing}
	
We evaluated the compression force $F_c=p_c A$ produced using an FFMS prototype with 3 channels (Fig. \ref{fig:testing_Force}H), where $p_c$ was the compressive pressure.  The actuator was wrapped around a cylinder in an isometric test fixture using a force sensor (ATI Nano17, ATI Industrial Automation). The contact area, $A$, was $209$ mm$^2$. We varied the applied fluid volume from 0 to 2 mL (Fig. \ref{fig:testing_Force}I).  This yielded compressive forces ranging from 0 to 10 N. These values were consistent with our predictions based on the range of axial forces, $F_{ext},$ produced by the same device: at fluid pressure $p=250$ kPa, we measured the the axial force to be $F_{ext}=13$ N. The model predicts a compressive pressure  $p_c = 44.5$ kPa for our text fixture configuration, which implies $F_c = p_c A = 9.3$ N, in close agreement (error $<$5\%) with our measurements (Fig. \ref{fig:testing_Force}K).  The results varied little for speeds below 5.0 Hz.

\subsubsection{Displacement}
\begin{figure*}[ht]
	\centering
	\includegraphics[width=170mm]{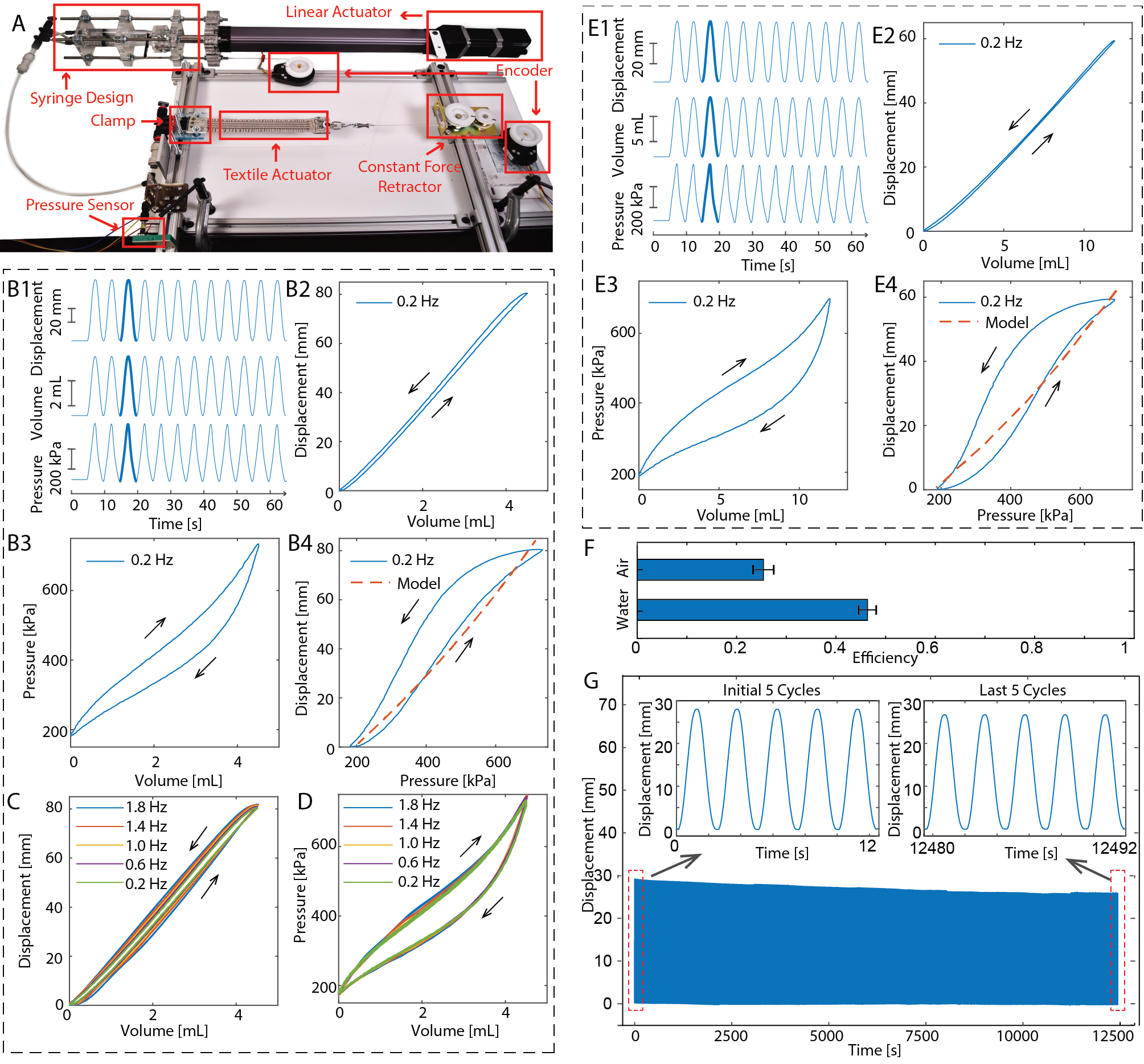} 
	\caption{
		Displacement and compression testing.  A. Apparatus for displacement testing. B-D. Results for the three-channel FFMS were consistent over repeated actuation.  Similar behavior was observed for different actuation speeds.  The pressure-displacement relationships were consistent with analytical predictions, despite  dynamic effects, including hysteresis (see text).  E1-E4. Results for the larger, 10-channel FFMS were similar to those for the smaller FFMS. F. Hydraulic operation with water was more efficient than with air.  Error bars: 95\% confidence intervals. G. Durability testing revealed consistent performance over 5000 actuation cycles. A 5\% reduction in displacement was observed after this period, which we attributed to initial actuator relaxation.
	}
	\label{fig:testing_Displacement} 
\end{figure*}

We evaluated the FFMS performance in displacement testing using a test configuration that was similar to the one we used in axial force testing (Fig. \ref{fig:testing_Displacement}A). For testing, a constant force retractor (Force 4.45N, model 61115A2, McMaster-Carr) was used to replace one of the isometric constraints in the fixture described above. An optical encoder was used to record the position for the distal end of FFMS. Displacement increased with  increases of fluid volume or fluid pressure.  The results revealed consistent displacement across repeated actuation (Fig.~\ref{fig:testing_Displacement}B1). As predicted, the relation between volume and displacement was almost perfectly linear, reflecting the incompressibility of the medium and radial constraint provided by the textile (Fig.~\ref{fig:testing_Displacement}B2). When the fluid volume reached 4.5 mL, the FFMS attained a length increase of 70\% (from 122.4 to 207.4 mm). In this case, the volume-pressure and pressure-displacement relationships exhibited greater hysteresis.  We attributed this to the  deformation of the nonlinear elastic materials at large displacements or high fluid volumes (Fig.~\ref{fig:testing_Displacement}B3-B4).  The analytical model correctly predicted the observed ranges of displacement, but because the model was quasi-static, it did not capture the hysteresis.  We plan to develop a dynamic model that can capture such effects in future work.	We also observed that the performance of FFMS remained consistent at higher driving frequencies. The results for the larger FFMS with 10-channels were similar to those that we observed for the smaller FFMS (see Fig. \ref{fig:testing_Displacement}E1-E4).

\subsubsection{Efficiency}
We computed the energetic efficiency as the ratio of the input energy and the mechanical work over one working cycle.  We measured this for a single channel FFMS with different working fluids, comprising water (hydraulic mode) and air (pneumatic mode). The tested FFMS had a rest length of 16 mm,  width of 3.7 mm, latex tube outer diameter of 6.4 mm, and tube inner diameter of 3.2 mm. The FFMS lifted a load of mass $m$ against gravity at speed of $v=0.1$ mm/s to height $h$.  Input work, $W_{in}$,  was computed as the sum of (positive) mechanical work, $W_+,$ performed when extending the actuator, and (negative) work, $W_-,$ performed when withdrawing the fluid (during lifting), yielding $W_{in}=W_+ + W_-$.  The energy efficiency, $R$, was $R = U_{g}/W_{in},$ where $U_g = mgh$ was the output work, or change in potential energy of the mass over one working cycle.  The  results were averaged over five working cycles in which the actuator returned to its initial state after displacing the mass.

The actuator efficiency was higher in hydraulic operation than in pneumatic operation  (efficiency $R=0.46$ vs.\ $0.25$, see Figure \ref{fig:testing_Displacement}F). This can be explained by the additional thermodynamic losses arising from the compression of air.  Other losses included those due to fluid viscosity, friction, and thermoelastic heating of the elastomer. Prior authors\cite{sridar2016hydro} computed the efficiency of a soft fluidic actuator by considering  the work $W_-$ perormed when withdrawing the fluid as an output of the system, yielding $R=(U_g + W_-) / W_+$. However, such a calculation leads to erroneous results, since a system can be designed to make the efficiency arbitrarily close to 1 by adding and removing an increment of fluidic energy at the input without any additional production of useful work (for example, this can be achieved by adding a reservoir at the input).  Applying this method to our actuator, we obtained a (erroneous) higher efficiency of $R=0.83$ in hydraulic operation.

\subsubsection{Durability}
We evaluated the durability of FFMS (three-channel prototype (Fig.~\ref{fig:testing_Displacement}G)) by actuating it over 5000 cycles to a maximum amplitude of 28 mm. The behavior was similar throughout testing. A  5\% reduction in displacement was observed after this testing cycle. We attributed this to initial relaxation of elastic tube and fabric material. The relatively consistent performance may be attributed, in part, to the operation of the tube within the elastic regime, which resulted in little plastic yielding.

\subsubsection{Failure Modes}

\begin{figure*}[ht]
	\centering
	\includegraphics[width=83mm]{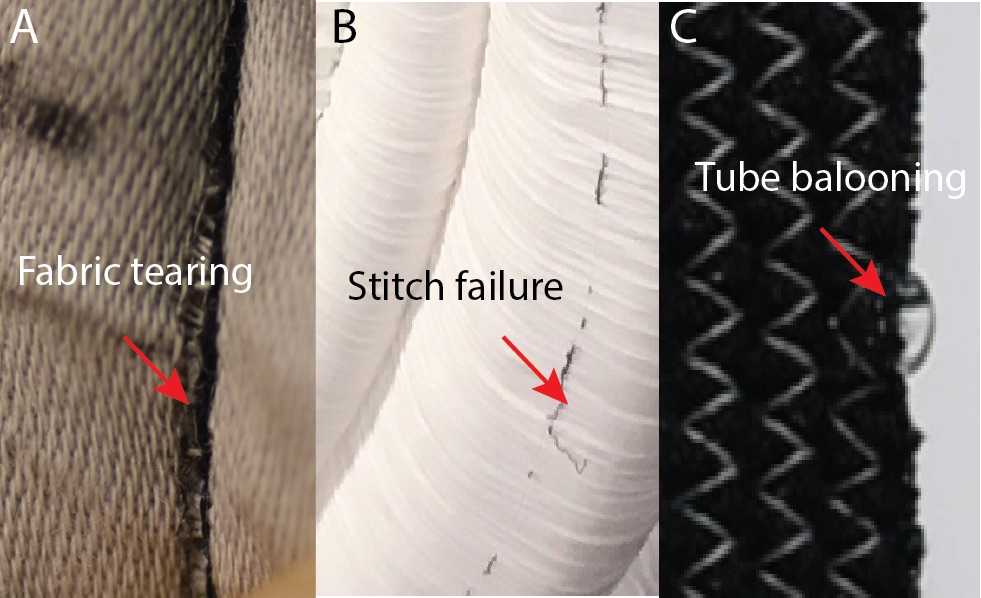}
	\caption{
		Examples illustrating failure modes. A. Fabric tearing at stitch locations. B. Stitching failure between fabric conduits. C. Tube ballooning between stitch locations.
	}
	\label{fig:failure} 
\end{figure*}

Actuator failure can arise due to improper selection of material, assembly, or to the operation regime.  Here, we highlight three typical failure modes observed in our prototypes during the experiments (Fig.\ \ref{fig:failure}). As the fluid pressure increases,  the elastic tubes, enclosed fabrics, and stitches are subjected to the increase of stresses. This can result in fabric tearing, stitch failure, or ballooning of the elastic tube. Fabric tearing can be minimized through the use of high-strength or dense fabrics, ideally with thread counts exceeding 300. 
One failure mode was associated with large cross section elastic tube. For fixed fluid pressure, $p$, the force per unit length, $t_s,$ exerted by the stitch is proportional to $p r_o,$ where $r_o$ is the tube outer radius.  As the tube radius is increased, stitch failure will occur when $t_s$ exceeds a critical value.  Such stitch failures may be minimized through the use of high strength threads, composed of materials such as polyester or Poly-paraphenylene terephthalamide (Kevlar), through multiple (double or triple) stitching, and through the use of  stronger fabrics.  Another failure mode arose from insufficient radial constraints on the elastic tube, which yielded radial expansion or ballooning of the tube.  We observed this to occur due to imperfections in side stitching, due to  gaps between cross stitches, or due to the use of four-way stretchable fabrics.  To mitigate such failures, consistent stitching should be used, four-way stretch fabric should be avoided, and, where two-way stretch fabrics are used, zig-zag configurations with smaller stitch distances (relative to the tube diameter) should be used.

\section{FFMS Embodiments and Applications}

FFMS actuators are versatile, and capable of actuating a variety of bodies or structures of different scales. They can be designed to realize multiple modes of actuation that are suited to applications in robotics, healthcare, and wearable technologies, as we illustrate below. 

\label{applications}

\subsection{Scalability}
\begin{figure*}[ht]
	\centering
	\includegraphics[width=170mm]{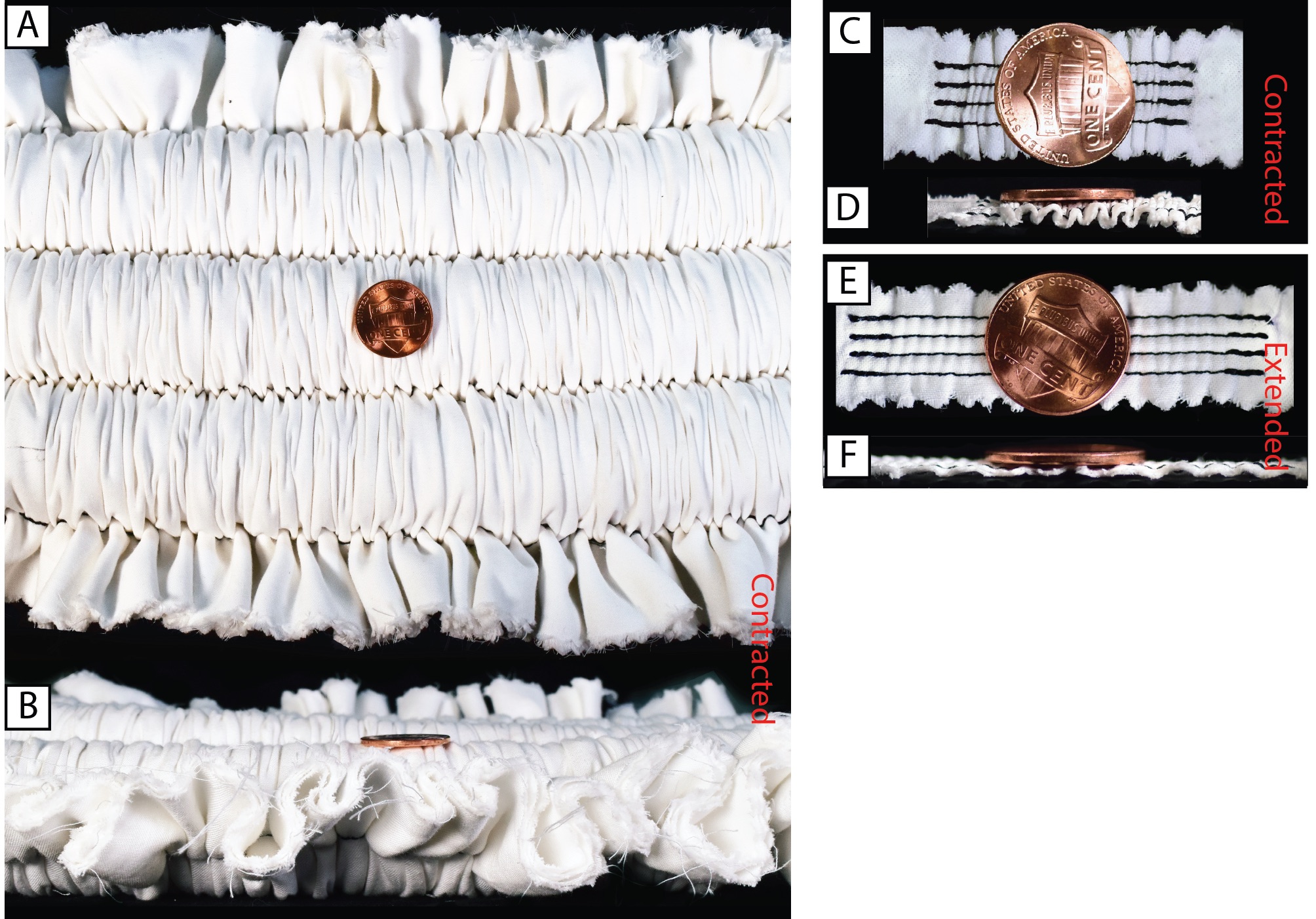} 
	\caption{
		FFMS actuators are readily scaled to small and large sizes.  (A,B) A large example, consisting of a 34.0 mm thickness FFMS  (A: Top view, B: Side view), shown  in contracted  state, was sufficient to lift 15 kg (Fig.~\ref{fig:application}).  (C-F) A small example, in contracted (C,D) and extended (E,F) states; the thickness when extended is 1.0 mm.  The large and small actuators were fabricated using the same general process.  A penny is used to illustrate the relative scale.
	}
	\label{fig:scalability} 
\end{figure*}

FFMS actuators can be used to realize devices of different sizes, ranging from millimeter to meter scale devices. The dimensional parameters of the FFMS include the lengths, $L,$ of the elastic tubes, their inner and outer radii, $r_i$ and $r_o$, and number, $N,$ of elastic tubes. These parameters  determine the stitched conduit width, fabric length,  and fabric width.  For fixed material, tube configuration, and strain, the maximum force is determined by $r_o^2-r_i^2$. The maximum elongation, $\Delta L,$ is proportioned to $L$.  The required maximum pressure is scale-invariant (Eq. \ref{eq:pmax}).  Commercially available elastic tubes can be used with widely varying radii,   $r_i, r_o$, ranging from less than 1 mm to greater than 30 mm.  We fabricated several prototypes to explore the scaling of FFMS actuators (Fig.\ \ref{fig:scalability}) for different applications (see Fig. \ref{fig:application}). Small, low-profile FFMS may be used in miniature biomedical or wearable devices, while large FFMS can be employed in higher force applications, such as orthotics or soft robotic construction machines.

\subsection{Multimodal Actuation and Shape Change}
\begin{figure*}[ht]
	\centering
	\includegraphics[width=170mm]{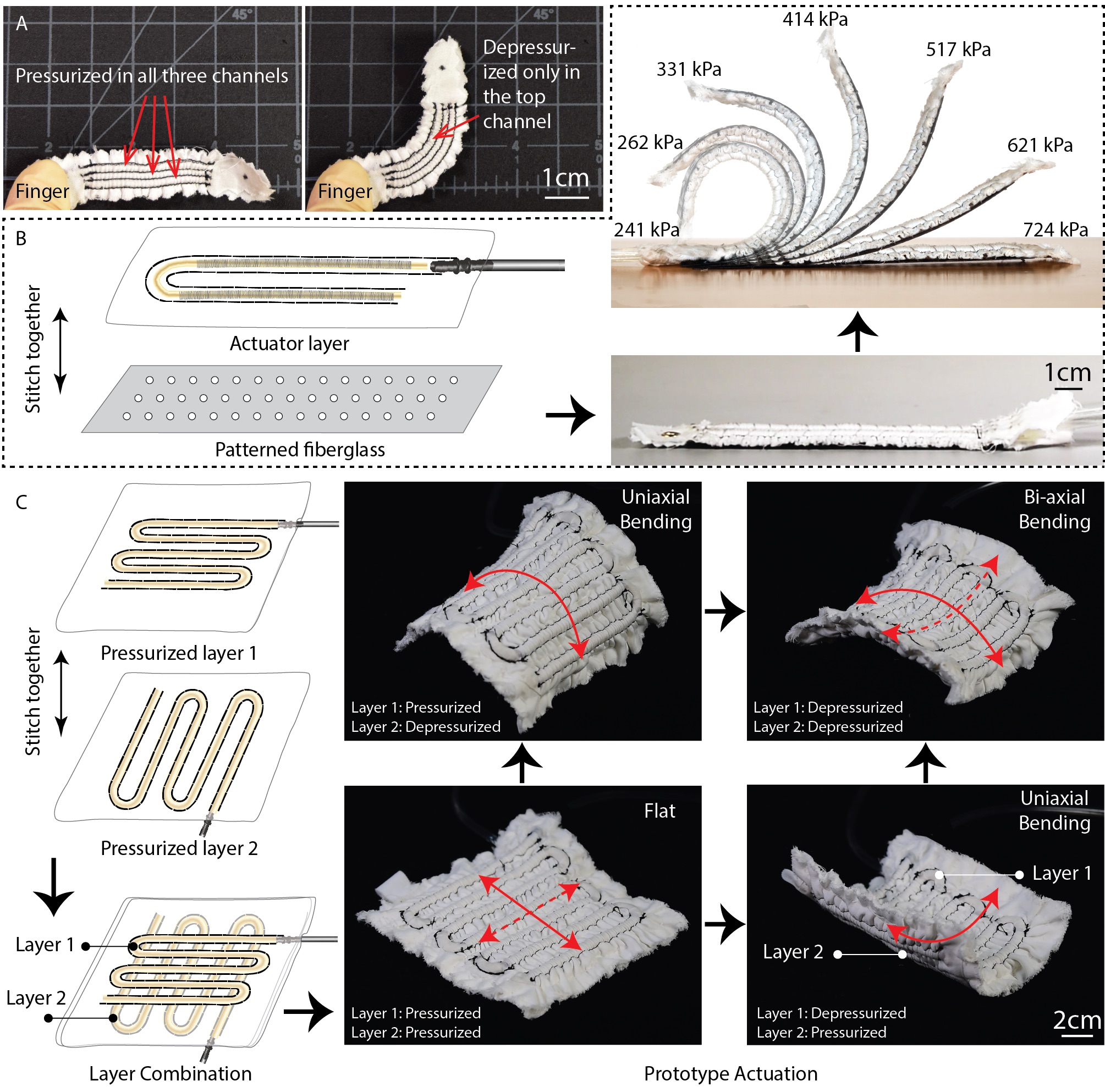}
	\caption{
		Composite or multi-actuated FFMS can realize dynamic, multimodal bending motion or shape change. (A)  In-plane rotation realized by differential pressurization of multiple fluid channels. (B)  Out-of-plane bending is realized by combining an FFMS with a second, passive layer with specified bending stiffness. (C) Biaxial bending is realized via a composite of two, orthogonally oriented FFMS sheet actuators.
	}
	\label{fig:potato_surfaceactuation} 
\end{figure*}

Composite or multi-actuated FFMS can yield dynamic, multimodal motion or shape change.   Here, we show how FFMS actuators can be used to realize in-plane rotation, out-of-plane bending, and biaxial bending motion (Fig.\ \ref{fig:potato_surfaceactuation}). 

In-plane rotation may be realized by differentially driving multiple fluid channels to steer an actuator in different extension modes.  Such motions may be used in soft biomedical devices \cite{berthet20182}.  In one embodiment, a 3 cm FFMS is driven via 3 elastic tubes that are independently controlled.  When one of three lateral tubes is depressurized, a large-amplitude  planar rotation can be achieved, yielding turning angles approaching 90 degrees (Fig.~\ref{fig:potato_surfaceactuation}A) due to differential elongation in the three tubes.  

out-of-plane bending motion can be achieved via the FFMS actuators. When a  passive layer of fiberglass (with specified bending stiffness) is combined with the FFMS, continuous out-of-plane bending can be generated. In one embodiment, a fiberglass sheet (thickness 0.38 mm) is pre-patterned via laser engraving and stitched to the FFMS (Fig.\ \ref{fig:potato_surfaceactuation}B). Pressurizing the FFMS yields large-amplitude bending, exceeding 180 degrees. Selecting the bending stiffness of the passive layer makes it possible to tune the actuator stiffness and generated torques. Such a configuration may be used in wearable or soft robotics applications \cite{Noritsugu2004medical}.

If two FFMS layers are oriented in complementary directions, biaxial bending, or 3D shape change, may be generated. To demonstrate this capability, we stitched two 7-channel FFMS layers together in orthogonal orientations (Fig.\ \ref{fig:potato_surfaceactuation}C). The whole composite structure remained flat when the applied fluid pressure was equal in both FFMS layers.  When the applied fluid pressures was unequal in each layer, bending motion was initiated about each of two orthogonal in-plane axes.  Biaxial bending yielded 3D or hyperboloid shapes. Such a biaxial FFMS evokes the transversus and rectus abdominis muscles in the human abdomen \cite{cutter1852treatise}, which enable the bending of the thorax or compression of the abdominal interior.

\subsection{Device Configurations and Applications}
\label{Device Configurations and Applications}
\begin{figure*}[ht]
		\centering
		\includegraphics[width=167mm]{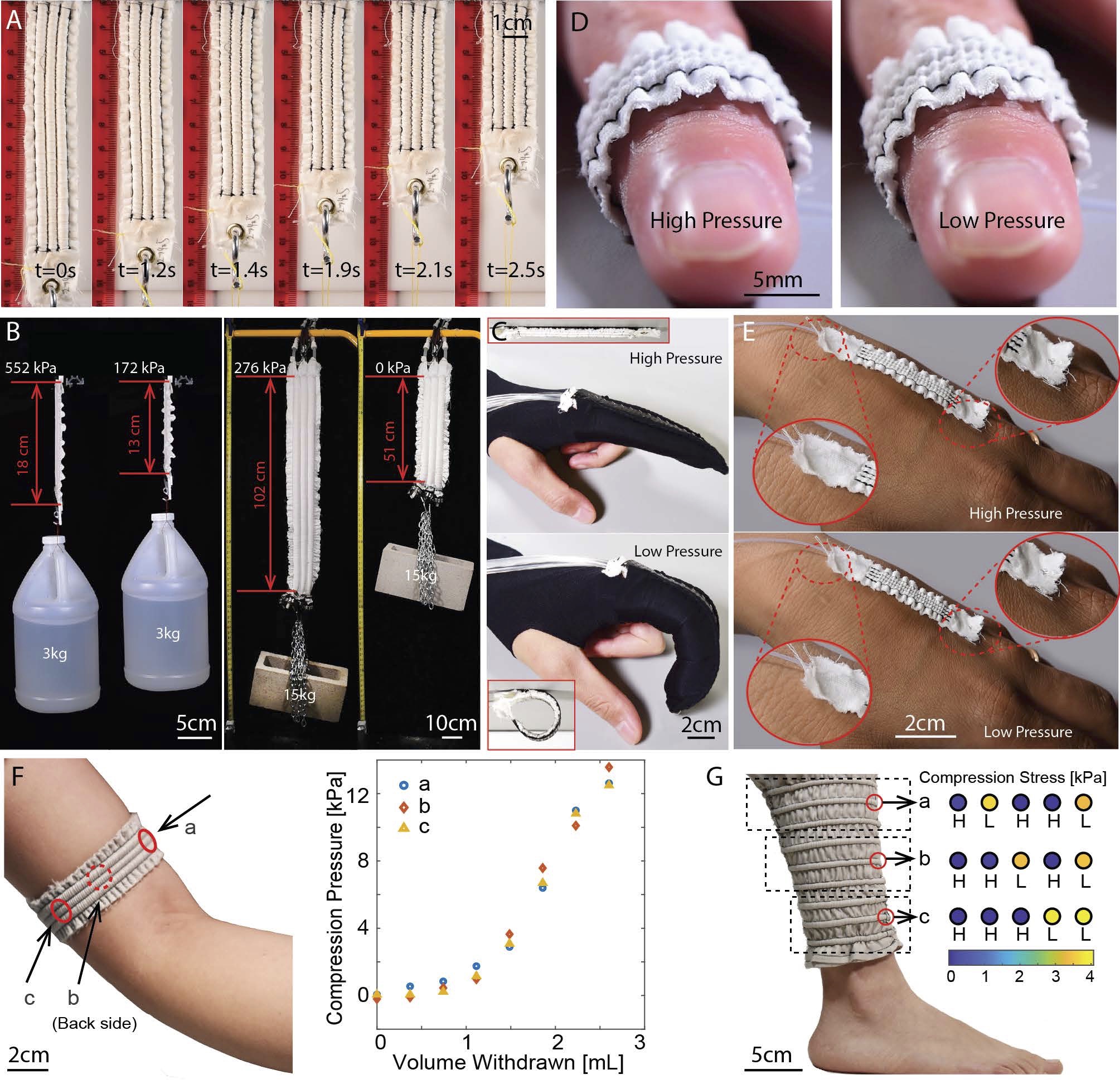}
\vspace{-2mm}
		\caption{
			Demonstrations. (A,D,E) Miniature soft actuators for linear motion control or compression, capable of (A) lifting a small mass, (D) compressing small tissue areas, or (E) providing tactile feedback via skin stretch.  Inset: Skin stretch was easily perceived. (B) FFMS actuators can perform large mechanical work.  A 10-channel device lifts a 3 kg mass. A 4-channel structure (inner tube radius 2 cm)  lifts a 15 kg cinder block and chain at pressures less than 276 kPa. (F,G) FFMS can be used for compression garments for healthcare, training, and haptics. (F) A compression band yields uniform pressure on the upper limb,  easily matching pressures provided by blood pressure cuffs.  (G) A compression garment for the lower limb comprises three independently addressable sections (dashed boxes).  Pressure variations can yield peristaltic motions suitable for undulatory massage  in therapies for lymphatic and blood circulation  \cite{duffield2016effects}, such as lymphedema  \cite{brennan1998overview} or venous closure \cite{partsch2005calf}.}
		\label{fig:application} 
	\end{figure*}

FFMS sheets may be used to realize motion or provide forces in a variety of applications, including soft robotic motion control, wearable actuators, assistive devices, and compression garments. We realized several examples (Fig.~\ref{fig:application} and Supplementary Video). Miniature FFMS can be used to produce linear motion, to provide skin stretch haptic feedback, or  to provide compression forces to small body parts, such as a finger (Fig.~\ref{fig:application}D,E). 
We also fabricated a 3-channel FFMS that can contract from 10.5 mm in to 5 mm in length, lifting a mass of 500 g, while producing engineering strains up to 110\% (Fig.\  \ref{fig:application}A). Other compact  devices can be used to apply constriction to small body parts (Fig. \ref{fig:application}D, showing an FFMS of 1 mm thickness). They can also be used to realize wearable devices for providing haptic feedback via skin-stretch, yielding highly palpable tactile sensations  (Fig.  \ref{fig:application}E).

Larger scale FFMS actuators can perform greater mechanical work. For example, we fabricated a  10-channel device that can lift a 3 kg mass  (Fig.\  \ref{fig:testing_Force}C). This  was 115 times higher than the actuator  mass  of 26 g (Fig.\  \ref{fig:application}B, left). We realized a  larger FFMS actuator with 3 tubes (inner latex tube of diameter  25 mm, outer diameter of 32 mm), which lifted a 15 kg mass at fluid pressures less than 276 kPa (Fig.\  \ref{fig:application}B, right). This demonstrates how FFMS actuators can be used to perform significant mechanical work. 

FFMS actuators  hold promise for wearable applications such as assistive devices.  Here, we employed a  configuration similar to the one we used for out-of-plane bending (Fig. \ref{fig:potato_surfaceactuation}C) in order to provide assistive flexion forces for two fingers during grasping (Fig.~\ref{fig:application}C). In this application, we pre-tensioned the fiberglass sheet layers such that the device performed flexion when pressurized and extension  when depressurized  (Fig.~\ref{fig:application}C, bottom inset).  The  force was sufficient to open and close the hand for grasping. Such a device is useful for assisting conditions such as stroke that can often lead to chronic flexion of the fingers, preventing grasping and adversely affecting many activities of daily living \cite{Noritsugu2004medical,polygerinos2015soft}.

FFMS devices can also be used to provide compression to larger areas of the extremities.  Such compressive forces are of interest for haptic feedback \cite{delazio2018force}, for   preventative compression therapy in deep vein thrombosis \cite{mazzone2002physical},  for  musculoskeletal recovery via blood flow restriction therapy \cite{loenneke2012low,lowery2014practical},  for  lymphatic or cardiovascular circulatory conditions \cite{king2012compression,williamson1994reflex}, or other biomedical devices  \cite{payne2018force,duffield2016effects, brennan1998overview,partsch2005calf}. We created an upper limb compression device based on an FFMS arm band and measured the resulting compressive pressures using thin-film force sensors (FlexiForce A201, Tekscan Inc.).  Withdrawing up to 3 mL of fluid from the band yielded uniform compression of up to 12 kPa, similar to pressures provided by blood pressure cuffs.  
We also demonstrate a wearable compression garment for the lower limb (Fig. \ref{fig:application}G). This device is constructed from 3 elastic tubes arranged in three independently controlled sections (Fig. \ref{fig:application}G, dashed boxes).  Supplying different fluid pressures to different sections yields varied compressive pressures across the limb.  This can be used to provide bulk compression  to the limb, which is useful in recovery from some injuries or in disease treatment. It can also be used for providing dynamic peristaltic motions that can be used for undulatory massage to augment lymph and blood flow. Such devices can meet the needs for pressure garment therapy \cite{duffield2016effects}, lymphedema treatments \cite{brennan1998overview}, or venous closure treatments \cite{partsch2005calf}.

	\section{Conclusions}
This paper presents a new family of soft actuators that we refer to as Fluidic Fabric Muscle Sheets, inspired by sheet-like biological muscles.  These devices comprise fabric layers with integrated hydraulic transmissions formed from arrays of hollow elastic tubes routed in patterned fabric conduits. We demonstrate how to design and fabricate these devices using facile methods that build on apparel engineering techniques including computerized sewing processes.  As we demonstrated, these devices are stretchable, conformable, safe, efficient, and scalable.  They are applicable to small, millimeter-scale actuators, and large meter-scale devices, and can yield forces exceeding 150 N, more than 115 times their weight, with engineering strains greater than 100\%. Laboratory prototypes perform consistently in testing over thousands of cycles. Their performance can be predicted via simple mechanical modeling, aiding design.  As we show, such FFMS actuators hold promise for applications in soft robotic motion control and for wearable devices for haptics, healthcare, and assistive technologies. The compressions they can produce meet requirements for several healthcare applications.

These results also point to several areas that are promising for future investigation. The fabrication methods we describe are simple and flexible, but further research is needed in order to align them with potential manufacturing techniques. FFMS actuators prove capable of performance in axial actuation, compression, and multimodal actuation, where dynamic shapes or stresses are enabled by designed routings of fluidic channels.  New analytical and computational design methods would facilitate a larger variety of programmable distributions of forces and strains, and would enable greater control over such behaviors.  The performance of FFMS actuators depends on the fluidic power source that is used. We highlighted advantages of hydraulic operation. Further research is needed on compact hydraulic power sources. The analytical model we present was effective for predicting actuator performance, but the model is quasi-static.  In future work, we plan to extend this approach to account for dynamics, aiding precise real-time control. Our devices employ open-loop control strategies, and the performance of these devices would be further improved through the use of closed-loop controllers relying on fluidic or strain sensors. Intrinsic mechanical or physiological sensors would enable further applications. The demonstration cases we present highlight potential applications in wearable devices for human-computer interaction and virtual reality. We anticipate investigating these in future work. Our demonstrations also point to a range of potential biomedical applications for assistive and therapeutic devices. These merit further research. We have highlighted applications of FFMS actuators in several forms of wearable devices.  Our design and fabrication methods are amenable to realizing integrated garments that may be applied to larger body areas, including the realization of whole-body actuated suits or soft exoskeletons, which  could greatly aid applications in haptic virtual reality, human space exploration, and rehabilitation.  We intend to explore such garments and applications in future work. 
	
	\section*{Symbols Used}
	\begin{table}[h]
		\centering % used for centering table
		\begin{tabular}{ll} 
			\hline
			Symbol & Description\\  [0.5ex]  
			\hline 
            $F_{fluid}$ & Axial force produced via fluid pressure \\ 
            $F_{ext}$ & Axial external load force \\
            $F_{el}$ & Axial elastic force of  tubing \\
            $F_{fab}$ & Axial force due to fabric \\
            $F_{d}$ & Dissipative forces including viscosity and friction \\
            $F_{d,hyd}$ & Hydrodynamic flow resistance force \\
            $F_{d,dry}$ & Dry friction force at tube-fabric interface \\
            $\tau$ & Shear stress of fluid at  inner wall of  tube \\          
            $\mathcal{A}$ & Contact area between fluid and  tube \\
            $\mu$ & Kinematic viscosity of fluid \\
            $\rho$ & Fluid density \\
            $\frac{ \partial u}{\partial y}$ & Rate of shear deformation of  fluid \\
            $r_i$ & Tube inner radius \\
            $L$ & Tube length\\
            $\mathcal{F_N}$ & Normal force between tube and fabric \\
            $\zeta$ & Friction coefficient \\
            $p$ & Fluid pressure \\
            $r_o$ & Tube outer radius \\
            $N$ & Number of elastic tubes \\
            $E$ & Tube elastic modulus  \\
            $\epsilon$ & Tube true strain \\
            $A_{tf}$ & Area of the tube-fabric interface \\
            $A_{tube}$ & Cross section area of a single tube \\
            $A_{fluid}$ & Cross section area of fluid inside a single tube \\
            $\delta L$ & Tube displacement \\
            $L_0$ & Initial tube length \\
            $h$ & Effective FFMS thickness \\
            $r_c$ & Radius of a cylindrical object \\
            $A_M$ & Effective axial cross section area of FFMS \\
			[1ex] % [1ex] adds vertical space
			\hline %inserts single line
		\end{tabular}
		\label{table:nonlin} % is used to refer this table in the text
	\end{table}

	\section*{Acknowledgement}
This work was supported by the US National Science Foundation under awards NSF-1628831, NSF-1623459, NSF-1751348 to Y.V. 
	
	\section*{Author Disclosure Statements}
	No competing financial interests exist. 
	
%	Immediately following the Acknowledgments section, include a section entitled ��Author Disclosure Statement.��
%	
%	All authors must disclose any associations that pose real or perceived conflicts of interest in connection with the manuscript. Authors should also disclose any financial interests that they may have in the company supporting the work. This statement should include appropriate information for EACH author, thereby representing that competing financial interests of all authors have been appropriately disclosed according to the policy of the Journal. It is important that all conflicts of interest, whether they are perceived, potential, or actual be disclosed. This information will remain confidential while the paper is being reviewed and will not influence the editorial decision. Please see the Uniform Requirements for Manuscripts Submitted to Biomedical Journals for further guidance. If no conflicts exist, the authors must state “No competing financial interests exist."
	
  \bibliographystyle{aiaa}

	\bibliography{zhu2019fabric.bib}

\end{document}